\documentclass[10pt,leqno]{amsart}
\usepackage{graphicx}
\pdfoutput=1
\usepackage{indentfirst,csquotes}

\usepackage{amsmath,amssymb,amsfonts}
\usepackage{algorithmic}
\usepackage{graphicx}
\usepackage{textcomp}
\usepackage{xcolor}
\def\BibTeX{{\rm B\kern-.05em{\sc i\kern-.025em b}\kern-.08em
    T\kern-.1667em\lower.7ex\hbox{E}\kern-.125emX}}

\usepackage[ruled,vlined]{algorithm2e}
\usepackage{booktabs}
\usepackage[table]{xcolor}
    
\begin{document}

\title[(Weighted) Adaptive Radius Near Neighbor Search]{(Weighted) Adaptive Radius Near Neighbor Search: Evaluation for WiFi Fingerprint-based Positioning}
\author[Le et al.]{Khang Le$^1$, Joaquín Torres-Sospedra$^2$, Philipp Müller$^3$}

\let\thefootnote\relax
\footnotetext{$^1$Faculty of Information Technology and Communication Sciences, Tampere University, Tampere, Finland, ORCID: 0009-0000-8327-4741

$^2$Department of Computer Science, University of Valencia,\textit{\% ValgrAI}, Valencia, Spain \\
ORCID: 0000-0003-4338-4334

$^3$Faculty of Medicine and Health Technology, Tampere University, Tampere, Finland, ORCID: 0000-0003-4314-7339\\

Version: 17 April 2026

This work has been submitted to the IEEE for possible publication. Copyright may be transferred without notice, after which this version may no longer be accessible.

P. Müller acknowledges funding from the Research Council of Finland under decision no. 360768. J. Torres-Sospedra acknowledges funding from Generalitat Valenciana (CIDEXG/2023/17, Conselleria d’Educació, Universitats i Ocupació).}

\maketitle

\begin{abstract}
Fixed Radius Near Neighbor (FRNN) search is an alternative to the widely used $k$ Nearest Neighbors ($k$NN) search. Unlike $k$NN, FRNN determines a label or an estimate for a test sample based on all training samples within a predefined distance. While this approach is beneficial in certain scenarios, assuming a fixed maximum distance for all training samples can decrease the accuracy of the FRNN. Therefore, in this paper we propose the Adaptive Radius Near Neighbor (ARNN) and the Weighted ARNN (WARNN), which employ adaptive distances and in latter case weights. All three methods are compared to $k$NN and twelve of its variants for a regression problem, namely WiFi fingerprinting indoor positioning, using 22 different datasets to provide a comprehensive analysis. While the performances of the tested FRNN and ARNN versions were amongst the worse, three of the four best methods in the test were WARNN versions, indicating that using weights together with adaptive distances achieves performance comparable or even better than $k$NN variants.
\end{abstract}


\section{Introduction}

In 1951, Fix and Hodges~\cite{Fix1989} introduced a non-parametric supervised learning method, named $k$ Nearest Neighbors ($k$NN), that is used until this day in numerous applications and for both classification and regression tasks~\cite{Sospedra2023}. The $k$NN remains one of the most suitable approaches for fingerprint-based positioning systems, particularly in technologies such as WiFi, Bluetooth Low Energy, and Visible Light Communication. In order to remain competitive with alternative machine learning techniques, which avoid exhaustive comparisons with the entire fingerprint map, resulting in faster inference, various optimization strategies have been proposed to accelerate $k$NN inference, ensuring that $k$NN continues to be a relevant and competitive approach.

The Fixed Radius Near Neighbor (FRNN) search has been around for nearly as long as the $k$NN, with~\cite{Bentley1975} attributing the origin of the approach to a paper by Levinthal from 1966~\cite{Levinthal1966}, but is markedly less used than the $k$NN. The basic principle behind the FRNN is to label a test sample based on the labels of all training samples within a given distance, which is used as similarity measure, from the test sample. Hence, the number of training samples for different test samples may vary. This is in contrast to the $k$NN, where the number of closest  training samples is fixed to $k$ for any test sample. Both methods have their advantages and disadvantages. Determining the label of a test sample based on a fixed number of $k$ training samples ensures that only samples most similar to the test sample are used. However, the $k$NN operates with relative similarities. Even if the test sample and its $k$ closest training samples show poor similarity, the $k$NN will still yield a label for the test sample. This can result in misclassifications, for example, if the test sample is from a class for which no labeled training samples are available. In such scenario the FRNN will return information that the test sample cannot be labelled, which is arguably more desirable than obtaining a misclassification. The drawback of the FRNN is that finding a radius yielding high accuracy while providing labels for (almost) all test samples is cumbersome, because training samples may be unevenly distributed across different areas.

To address these shortcomings various modifications of $k$NN and FRNN have been proposed (see e.g.~\cite{Sospedra2023,Uddin2022}) to improve their accuracy, with the majority focuses on the $k$NN. However, Torres-Sospedra et al.~\cite{Sospedra2023} demonstrated that none of the proposed improved $k$NN variants consistently outperformed all other variants when being applied to 69 fingerprint datasets dealing with indoor positioning based on WiFi, BLE, and hybrid signals. Instead, for different datasets different $k$NN variants emerged as the most accurate variants.

Motivated by their work and by the observation that different samples benefit from different radii, we propose two modifications of the FRNN with adaptive radii named the Adaptive Radius Near Neighbor (ARNN) and the Weighted Adaptive Radius Near Neighbor (WARNN). We compare their performance with FRNN and the $k$NN variants investigated in~\cite{Sospedra2023} for indoor positioning using 22 WiFi fingerprint datasets.

This paper demonstrates that the performance of the traditional FRNN is inferior to that of the $k$NN variants, while the newly proposed ARNN performed on par. It, furthermore, shows that the second newly proposed method, the WARNN, can even outperform all tested $k$NN variants.


\section{Related work}\label{sec:RelWorks}

Since the first introduction of $k$NN in indoor positioning in 2000~\cite{Bahl2000}, a significant number of its variants has been developed with the aim of improving positioning accuracy by utilizing different techniques. To compare their performance, \cite{Sospedra2023} examined 13 $k$NN implementations on 69 datasets. The results showed that adaptive weighted methods such as Adaptive Weighted $k$NN~\cite{Liu2022} and Self-Adaptive Weighted $k$NN~\cite{Hu2019} performed best amongst the evaluated models. However, \cite{Sospedra2023} demonstrated that no single $k$NN variant outperformed all other variants for all datasets.

FRNN was used for indoor positioning based on ion mobility spectrometry measurements in~\cite{Muller2023}. In the FRNN a predefined radius is used to determine the neighbors of a test sample instead of a fixed value of $k$. \cite{Muller2023} demonstrated that the FRNN can outperform the $k$NN when a suitable radius is selected. However, this advantage came at the cost of some test samples being not classified due to the absence of training samples within the predefined radius. In \cite{Muller2023} FRNN and $k$NN were evaluated on a single dataset, thus limiting the robustness of the conclusions. Therefore, this article examines the performance of both FRNN and the $k$NN variants tested in~\cite{Sospedra2023} on 22 WiFi datasets.


\section{Radius-based Near Neighbors methods}\label{sec:RNNs}

\subsection{Fixed Radius Near Neighbor}

While $k$NN variants require the number of closest neighbors $k$ as input, the Fixed Radius Near Neighbor requires a maximum radius $r_{\text{max}}$ as input parameter. This radius is a measure of the similarity between a test sample $\mathbf{x} = [ x_{1}, \ldots , x_{n} ] \in \mathbb{R}^n$ and any training sample $\mathbf{y}_i$, $i={1,..,N}$. Instead of searching for the $k$ training samples with the highest similarity (i.e. smallest distance) to $\mathbf{x}$ as for $k$NN-type methods, the FRNN searches for all training samples that are within $r_{\text{max}}$ to the test sample and determines a label for the test sample based on the training samples within $r_{\text{max}}$. Thus, $r_{\text{max}}$ can be interpreted as minimum similarity required for accepting training samples as being predictive of the label of the test sample $\mathbf{x}$. Therefore, $k$ is variable and an output parameter, if desired, of the FRNN.

The FRNN has some theoretic advantages over $k$NN-type methods. First, it does not require finding a suitable or even an optimal $k$. Second, if the test sample is from a class for which no training samples are available in the training set $\mathbf{Y}$, then any $k$NN variant will misclassify the test sample $\mathbf{x}$ as belonging to one of the classes present in $\mathbf{Y}$. The FRNN, in contrast, could return in such scenario information that no label could be provided because none of the training samples was within $r_{\text{max}}$ of the test sample. However, this cannot be guaranteed and the FRNN might still yield a misclassification (see e.g. the test results in~\cite{Muller2023}). Furthermore, choosing a suitable $r_{\text{max}}$ can be challenging and cumbersome. This is explainable by the fact that in most real-world scenario training samples are not uniformly distributed, meaning that training sample density varies in different regions. Therefore, in this paper, two Radius Near Neighbor (RNN) variants inspired by adaptive $k$NN and weighted $k$NN are proposed.

\subsection{Adaptive Radius Near Neighbors}

The Adaptive $k$NN (A$k$NN) was proposed to resolve the challenge of selecting an appropriate $k$ \cite{Wettschereck1993,Sun2010}. Instead of relying on a fixed value, it dynamically determines an optimal $k$ for each training point by using a separate algorithm. Each test point then finds its nearest neighbor and inherits that $k$ value from that neighbor~\cite{Uddin2022}. 

Inspired by A$k$NN we propose the Adaptive Radius Near Neighbors (ARNN) as an alternative to the FRNN. Instead of requiring the user to define $r_{\text{max}}$, the ARNN determines a suitable $r_i$ for any training sample $\mathbf{y}_i$ in the training phase (see Algorithm~\ref{alg:arnn_warnn} for the pseudo-code). The algorithm determines for any $\mathbf{y}_i$ ($i={1, \ldots , N}$) its distances to the remaining training samples. It then tests for the $k$ ($k=\{K_{\text{min}},..,K_{\text{max}}\}$) closest neighbors if they would yield a correct label in case of a classification task or an accurate estimate in case of a regression task. At the end, it sets $r_i$ to the largest distance between the training sample and its $k$ neighbors for the largest $k$ that yields a correct label or accurate estimate. Instead of the largest distance other statistics such as mean or median values could be used, but will not be studied in this manuscript. If none of the tested $k$ values yields a correct label or accurate estimate, then $r_i$ is set to zero and $\mathbf{y}_i$ is likely an outlier.

\begin{algorithm}
\caption{Training phase for ARNN and WARNN.}
\label{alg:arnn_warnn}
\SetAlgoLined
\textbf{Input:} training samples $\mathbf{Y} = [\mathbf{y}_1, \ldots , \mathbf{y}_N] \in \mathbb{R}^{n \times N}$ from $M$ classes, $K_{\text{min}}$, $K_{\text{max}}$, error threshold $\tau_{\epsilon}$ (for regression task)\\
\textbf{Output:} radii $\mathbf{r} = [r_1, \ldots , r_N]$\\
\For{$i \leftarrow 1$ \KwTo $N$}{
	Set $r_i \leftarrow 0$\\
	Calculate distances $d(\mathbf{y}_i,\mathbf{y}_j)$ for all  $j=1,\ldots,N$ with $j\neq i$\\
	Sort distances $d(\mathbf{y}_i,\mathbf{y}_j)$ in ascending order and store them in $\mathbf{d}_i$ and the corresponding $j$ in $\mathbf{I}_i$\\
\For{$K \leftarrow K_{\text{min}}$ \KwTo $K_{\text{max}}$}{
	Calculate a class label (classification task) or a numeric estimate (regression task) for $\mathbf{y}_i$ from $K$ training samples associated to the first $k$ indices in $\mathbf{I}_i$\\
	If the label is correct (classification task) or the estimate within $\tau_{\epsilon}$ (regression task) then set $r_i \leftarrow K$-th element in $\mathbf{d}_i$.}
}
\end{algorithm}

For determining a label or estimate for the test sample $\mathbf{x}$ (see Algorithm 2 for the pseudo-code), the ARNN determines a set $C$ of training samples for which the test sample is within the radii computed in the training phase. It then computes a label or an estimate based on the training samples in $C$. If $C$ is empty, meaning that $\mathbf{x}$ is not within the neighborhood of any training sample, the ARNN returns information that no label or estimate could be determined, indicating that the test sample is potentially an outlier or from a class that is not present in the training data.

\begin{algorithm}
\caption{Test phase for ARNN and WARNN.}
\SetAlgoLined
\textbf{Input:} training samples $\mathbf{Y} = [\mathbf{y}_1, \ldots , \mathbf{y}_N]$, radii $\mathbf{r} = [r_1, \ldots , r_N]$, test sample $\mathbf{x} = [ x_{1}, \ldots , x_{n} ] \in \mathbb{R}^n$\\
\textbf{Output:} Class label (classification task) or numeric estimate (regression task) for $\mathbf{x}$\\
Set $C \leftarrow \emptyset$\\
\For{$i \leftarrow 1$ \KwTo $N$}{
	Calculate distance $d(\mathbf{x}_i,\mathbf{y}_i)$\\
	If $d(\mathbf{x}_i,\mathbf{y}_i) \leq r_i$ then add $i$ to set $C$
}
\eIf{$C \neq \emptyset$}{
	Calculate a label or estimate based on the training samples with indices in $C$
}%
{   Return that no label (classification task) or estimate (regression task) could be determined due to lack of close neighbors
}
\end{algorithm}


\subsection{Weighted Adaptive Radius Near Neighbors}

One point of criticism for the $k$NN is that it assumes that all $k$ neighbors are equally informative when determining a label or estimate for a test sample. However, this assumption is often not fulfilled. For example, when determining a location estimate based on WiFi fingerprints, then training samples with fingerprints almost identical to the fingerprint at the test location most likely provide a better estimate than training samples with fingerprints only roughly the same as the test fingerprint. Therefore, an extension to $k$NN was introduced that assigns weights to the $k$ neighbors, known as Weighted $k$ Nearest Neighbors (W$k$NN)~\cite{Dudani1976}. It gives higher weights to closer neighbors and lower weights to distant neighbors, with weights summing up to one. In scenarios where closer neighbors are more representative of the test sample than neighbors further away, this strategy, in general, improves accuracy~\cite{Uddin2022}.

\begin{figure}[!hbt]
\centering
\includegraphics[width=.32\textwidth,clip=true, trim=0cm 0cm 0cm 0cm]{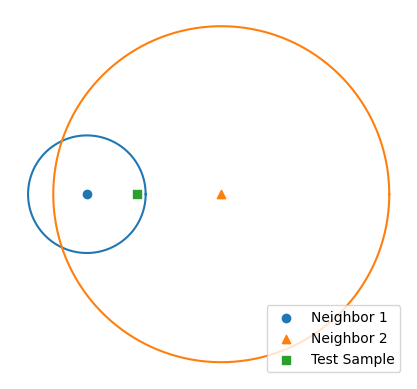}
\caption{
Illustration of challenge to choose suitable weights for the WARNN.
}
\label{fig:exWeightFct}
\end{figure}

Based on this intuition, we propose the use of weights in the ARNN and name it consistently the Weighted Adaptive Radius Near Neighbors (WARNN). In contrast to the W$k$NN, for the ARNN not only the distance between the test sample $\mathbf{x}$ and the training sample $\mathbf{y}_i$ matters, but also how close this distance is to $r_i$. Consider, a scenario as illustrated in Fig.~\ref{fig:exWeightFct}. The test sample is closer to neighbor 1 than to neighbor 2 in a two dimensional feature space. With the standard weighting function employed commonly in the W$k$NN, the first neighbor would be assigned a larger weight than the second. However, $\mathbf{x}$ is almost at the edge of the neighborhood defined by the radius of neighbor 1 but only half-way between neighbor 2 and the circle defined by its radius. This indicates that the test sample is weakly associated with neighbor 1 but strongly associated with neighbor 2.

Therefore, we propose an adaptive decay factor for the inverse distance weighting (IDW) function that reflects the distance between the test sample $\mathbf{x}$ and a neighbor $\mathbf{y}_i$ as well as the relative position of the test sample in the sphere of $\mathbf{y}_i$ defined by radius $\mathbf{r}_i$. The IDW function is defined by
\begin{equation}
w_i=\frac{1}{d(\mathbf{y}_i,\mathbf{x})^{\alpha}}, \label{weightfct}
\end{equation}
where $\alpha$ is the decay factor. As the distance $d(\mathbf{y}_i,\mathbf{x})$ increases the weight decreases, as in the IDW commonly used by W$k$NN methods. The decay factor $\alpha$ enables controlling the rate at which weights decrease. The higher the decay factor the faster the decay. No optimal decay factor has been reported in the literature and different studies have used values between 1 and 5~\cite{Bekele2003,Ping2004,Lloyd2005,Lu2008}. Therefore, we propose an adaptive decay factor defined by
\begin{equation}
\alpha=1+\frac{d(\mathbf{y}_i,\mathbf{x})}{r_i}, \label{decayfct}
\end{equation}
which ensures that $\alpha \in [1,2]$ and reflects the relative position of the test sample within the sphere around the training sample while ensuring that closer neighbors retain higher influence on the overall label or estimate respectively.


\subsection{Coverage ratio}

For all three RNN variants it is possible that no label or estimate is returned when the test sample lies outside the radii of all training sample. Although this property is desirable when analyzing samples from unknown classes, it becomes problematic for classification tasks when a large portion of test samples from classes present in the training samples are not classified. For regression tasks it may yield overly optimistic average estimation errors due to yielding only estimates for test samples most similar to any training sample.

Thus, to evaluate the performance of RNN methods one should also consider the ratio of test samples for which labels or estimates are provided. This ratio, which is herein referred to as coverage ratio, is defined as
\begin{equation}
\gamma=100\% \frac{h}{m}, \label{CR}
\end{equation}
where $h$ is the number of test samples receiving a label or estimate and $m$ is the total number of test samples ($h \leq m$).


\section{Empirical Analysis and Results}\label{sec:Appl}

\subsection{Experimental setup}

We applied FRNN, ARNN, and WARNN with two different weighting functions for determining 3-dimensional position estimates and compared them to the $k$NN and its variants analyzed in~\cite{Sospedra2023}.  Conversely to~\cite{Sospedra2023}, the evaluation includes a relevant subset of 22 diverse WiFi datasets widely used in the literature: DSI$n$~\cite{Moreira2020}, LIB$n$~\cite{Mendoza2018}, TUT$n$~\cite{Shreshta2013,Razavi2015,Cramariuc2016,Lohan2017,Richter2018,Lohan2020}, UJI$n$~\cite{Sospedra2014}, SOD$n$~\cite{Bi2022}. The implemented $k$NN variants and dataset descriptions are provided in~\cite{Sospedra2023supmat} for research reproducibility and replicability~\cite{Anagnostopoulos2021}.

The tested $k$NN variants included:
\begin{itemize}
\item $M_1$: $k$NN with optimal $k$ and unweighted centroid
\item $M_2$: $k$NN with optimal $k$ and weighted centroid (IDW)
\item $M_3$: $k$NN with optimal $k$ and weighted centroid (IDW\textsuperscript{2})
\item $M_4$: Adaptive W$k$NN~\cite{Liu2022} (AW$K$NN) with $k_{\text{max}}=51$
\item $M_5$: AW$K$NN with optimal $k_{\text{max}}$ based on W$k$NN
\item $M_6$: Self-Adaptive W$k$NN~\cite{Hu2019} (SAW$K$NN) with $k_{\text{max}}=51$
\item $M_7$: SAW$K$NN with optimal $k_{\text{max}}$ 
\item $M_8$: Spatial-Temporal Improved W$k$NN~\cite{Zou2017} (STIW$k$NN) with optimal $k$
\item $M_9$: Distance Weighted and Feature Weighted $k$NN~\cite{Liang2012} (DWFW$k$NN) with physical distance from~\cite{Liang2012}
\item $M_{10}$: DWFW$k$NN with physical distance from~\cite{Sospedra2023}
\item $M_{11}$: Adaptive Residual W$k$NN~\cite{Xu2020} (ARW$k$NN) with optimal $k$ and Cityblock distance 
\item $M_{12}$: ARW$k$NN with optimal $k$ and Min-Max distance 
\item $M_{13}$: ARW$k$NN with optimal $k$ and Clark distance 
\end{itemize}
$M_{1-3}$, $M_6$ and $M_7$ used Cityblock, $M_4$ and $M_5$ Cosine, and $M_8$ Euclidean distance metric. Used distance metrics and optimal $k$ values were determined by brute-force search. Tested distance metrics included Euclidean, Cityblock, Min-Max, Cosine, and Clark; tested $k$ values included $\{1,2, \ldots ,20,21,23, \ldots ,51\}$. Details on the used $k$NN variants are given in~\cite{Sospedra2023} and references therein.

In addition, the following RNN variants were tested
\begin{itemize}
\item $M_{14}$: FRNN with Euclidean distance
\item $M_{15}$: FRNN with Cityblock distance
\item $M_{16}$: FRNN with Cosine distance
\item $M_{17}$: ARNN with Euclidean distance
\item $M_{18}$: ARNN with Cityblock distance
\item $M_{19}$: ARNN with Cosine distance
\item $M_{20}$: WARNN with Euclidean and IDW (1) with $\alpha=2$
\item $M_{21}$: WARNN with Euclidean and IDW (1) with $\alpha$ defined by (2)
\item $M_{22}$: WARNN with Cityblock and IDW (1) with $\alpha=2$
\item $M_{23}$: WARNN with Cityblock and IDW (1) with $\alpha$ defined by (2)
\item $M_{24}$: WARNN with Cosine and IDW (1) with $\alpha=2$
\item $M_{25}$: WARNN with Cosine and IDW (1) with $\alpha$ defined by (2)
\end{itemize}
For all RNN variants a minimum coverage ratio of 90\% was required. Optimal $r_{\text{max}}$ values were determined by brute-force search for $M_{14-16}$ from the following sets: $\{60,62,  \ldots, 260 \}$ for $M_{14}$, $\{150, 152, \ldots, 300\} \cup \{305, 310, \ldots, 1000\} \cup \{1010, 1020, \ldots, 2350\}$ for $M_{15}$, and $\{ 1.0 \cdot10^{-2},1.2\cdot10^{-2}, \ldots, 9.8\cdot10^{-2} \} \cup \{ 0.100, 0.105, \ldots, 0.500 \}$ for $M_{16}$. In addition, for ARNN and WARNN $\tau_{\epsilon}=5$\,m was used, and $K_{\text{min}}=1$ and $K_{\text{max}}=\text{max} \left( K_{\text{min}}, \{  \left\lceil p\% \cdot N \right\rceil \right)$ were tested to limit the maximum number of neighbors per training sample to (approximately) $p$\% of the number of training samples. Only values from the set $p \in [0.1, 0.2, \ldots, 25.0] \cup [25, 26, \ldots ,40]\}$ ensuring a coverage ratio of at least 90\% were tested.

In the positioning task the three-dimensional (3D) position estimate $\hat{\mathbf{z}}=[\hat{z}_1, \hat{z}_2, \hat{z}_3]$ for test location $\mathbf{z}=[z_1, z_2, z_3]$ given the WiFi Received Signal Strength (RSS) fingerprint $\mathbf{y}$ at the test location was determined using the locations of the closest WiFi RSS fingerprints in the training dataset according to the 25 different methods described above. For algorithms incorporating weights, all weights were scaled to sum up to one. The 3D positioning error $e_{3\text{D}}$ for $\hat{\mathbf{z}}$ was calculated by
\begin{equation}
e_{3\text{D}} = \sqrt{ \left( \hat{z}_1 - z_1 \right)^2 + \left( \hat{z}_2 - z_2 \right)^2 + \left( \hat{z}_3 - z_3 \right)^2}. \label{3D_err}
\end{equation}


\subsection{Comparison of $k$NN and RNN variants}

Table~\ref{tab:3Derrors} reports the mean 3D positioning errors for each dataset across all methods $M_{1-25}$, as well as the average of mean 3D positioning errors computed over all 22 datasets. In addition, for $M_{14-25}$ coverage ratios for each dataset as well as the average coverage ratio over all 22 datasets are reported. 

\begin{table}[t]
\caption{
Dataset-wise mean 3D positioning errors and average 3D positioning errors for $k$NN and RNN implementations.}
\label{tab:3Derrors}
\centering
\scriptsize
\resizebox{\textwidth}{!}{%

\begin{tabular}{l|*{13}{r} |*{3}{r}|*{3}{r}|*{6}{r}}

\toprule
Dataset & $M_{1}$ & $M_{2}$ & $M_{3}$ & $M_{4}$ & $M_{5}$ & $M_{6}$ & $M_{7}$ & $M_{8}$ & $M_{9}$ & $M_{10}$ & $M_{11}$ & $M_{12}$ & $M_{13}$ & $M_{14}$ & $M_{15}$ & $M_{16}$ & $M_{17}$ & $M_{18}$ & $M_{19}$ & $M_{20}$ & $M_{21}$ & $M_{22}$ & $M_{23}$ & $M_{24}$ & $M_{25}$ \\
\midrule
DSI1  &4.09  &4.05  &4.03  &3.86  &3.83  &4.10  &4.03  &3.82  &\cellcolor{red!25}27.63  &4.14  &4.78  &3.91  &3.90  &5.14  &\cellcolor{orange!40}5.80  &4.06  &3.84  &4.06  &3.79  &3.74  &3.67  &3.74  &3.61  &\cellcolor{green!25}3.57  &3.67  \\
DSI2  &4.22  &4.17  &4.11  &3.85  &3.84  &4.05  &4.05  &3.83  &\cellcolor{red!25}27.88  &4.11  &4.90  &3.95  &3.93  &5.13  &\cellcolor{orange!40}5.75  &4.03  &3.98  &4.05  &4.02  &3.87  &3.75  &3.74  &\cellcolor{green!25}3.59  &3.83  &3.91  \\
LIB1 &2.44  &2.43  &2.42  &2.47  &2.44  &2.43  &\cellcolor{green!25}2.41  &2.44  &\cellcolor{red!25}4.11  &2.50  &2.47  &2.61  &2.90  &2.69  &2.75  &2.69  &2.87  &2.86  &\cellcolor{orange!40}2.91  &2.75  &2.71  &2.69  &2.66  &2.56  &2.71  \\
LIB2 &3.70  &3.70  &3.69  &\cellcolor{green!25}2.64  &\cellcolor{green!25}2.64  &3.66  &3.68  &2.68  &\cellcolor{red!25}5.59  &3.35  &\cellcolor{orange!40}4.20  &2.69  &3.25  &3.59  &3.57  &2.77  &3.17  &3.00  &3.21  &3.15  &3.16  &3.01  &2.99  &3.03  &3.06  \\
TUT1 &8.28  &8.20  &8.06  &6.21  &\cellcolor{green!25}6.16  &7.67  &8.21  &6.28  &\cellcolor{red!25}11.87  &7.70  &\cellcolor{orange!40}9.41  &6.93  &6.95  &7.90  &8.39  &6.81  &7.96  &7.15  &7.97  &7.57  &7.51  &6.90  &6.87  &6.81  &7.14  \\
TUT2 &11.23  &11.02  &10.85  &8.84  &8.83  &11.17  &11.17  &8.79  &\cellcolor{red!25}16.22  &9.98  &\cellcolor{orange!40}12.40  &9.72  &9.63  &11.17  &12.22  &8.80  &9.59  &10.09  &9.49  &9.53  &9.20  &10.22  &9.69  &\cellcolor{green!25}8.76  &9.03  \\
TUT3 &9.06  &8.65  &8.36  &8.33  &8.33  &8.35  &8.32  &8.24  &\cellcolor{red!25}20.55  &8.57  &9.16  &8.34  &8.39  &10.76  &\cellcolor{orange!40}13.47  &8.49  &8.45  &9.13  &8.49  &7.95  &\cellcolor{green!25}7.90  &8.38  &8.37  &8.24  &8.33  \\
TUT4 &5.68  &5.57  &5.52  &5.83  &5.83  &5.76  &5.58  &5.80  &\cellcolor{red!25}20.27  &5.81  &6.25  &5.68  &5.49  &7.73  &\cellcolor{orange!40}9.27  &6.32  &5.62  &5.92  &5.74  &5.32  &\cellcolor{green!25}5.28  &5.45  &5.50  &5.32  &5.57  \\
TUT5 &6.29  &6.23  &6.19  &\cellcolor{green!25}5.82  &5.93  &6.14  &6.07  &5.93  &\cellcolor{red!25}16.38  &6.39  &6.92  &6.11  &5.84  &6.77  &\cellcolor{orange!40}8.35  &6.27  &6.89  &6.47  &6.38  &6.57  &6.45  &6.19  &6.12  &6.18  &6.41  \\
TUT6 &1.87  &1.87  &1.87  &2.12  &2.11  &1.76  &1.75  &2.11  & \cellcolor{red!25}14.32 & 1.95 & 1.89 & 1.82 & 2.01 &4.05  &\cellcolor{orange!40}4.34  &3.04  &2.21  &2.21  &2.27  &1.92  &1.77  &1.84  &\cellcolor{green!25}1.66  &1.91  &2.24  \\
TUT7 &2.23  &2.23  &2.15  &2.53  &2.51  &2.10  &2.10  &2.51  & \cellcolor{red!25}14.34 & 2.23 & 2.24 & 2.11 & 2.24  &3.17  &\cellcolor{orange!40}3.39  &2.99  &2.41  &2.43  &2.57  &2.10  &1.98  &2.01  &\cellcolor{green!25}1.87  &2.18  &2.52  \\
UJI1 &8.62  &8.61  &8.60  &7.60  &7.58  &8.52  &8.46  &7.59  & \cellcolor{red!25}29.51 & 7.97 & 9.42 & 7.91 & 7.86  &10.35  &\cellcolor{orange!40}11.98  &7.77  &8.64  &8.36  &8.50  &8.14  &7.97  &7.70  &\cellcolor{green!25}7.46  &8.04  &8.16  \\
UJI2 & 6.35 &6.33  &6.31  &6.66  &6.62  &6.41  &6.31  &6.60  & \cellcolor{red!25}21.13 & \cellcolor{green!25}6.03 & 7.02 & 6.73 & 6.53  &7.91  &\cellcolor{orange!40}8.38  &7.41  &7.20  &7.51  &6.83  &6.93  &6.84  &6.98  &6.80  &6.50  &6.58  \\
SOD1 &2.65  &2.65  &2.64  &2.98  &2.84  &2.67  &2.64  &2.83  & \cellcolor{red!25}5.45 & 2.66 & 3.16 & 2.98 & \cellcolor{green!25}2.50  &3.41  &3.13  &\cellcolor{orange!40}3.73  &2.98  &2.77  &3.11  &2.93  &2.88  &2.72  &2.62  &3.05  &3.19  \\
SOD2 &1.72  &1.71  &1.71  &1.93  &1.99  &1.68  &1.68  &2.00 &\cellcolor{red!25}4.38 & 2.48 & 1.85 &1.65 & 1.92  &\cellcolor{orange!40}2.75  &2.41  &2.26  &2.04  &1.79  &1.89  &1.91  &1.86  &1.61  &\cellcolor{green!25}1.52  &1.77  &1.78  \\
SOD3 &1.80  &1.80  &1.79  &1.97  &1.98  &1.73  &1.79  &2.07  & \cellcolor{red!25}7.20 & 2.46 & 1.90 & 1.79 & 1.82  &\cellcolor{orange!40}3.07  &2.64  &2.54  &2.14  &1.82  &2.15  &2.04  &1.96  &1.69  &\cellcolor{green!25}1.58  &1.97  &2.08  \\
SOD4 &2.50  &2.50  &2.50  &4.09  &4.09  &\cellcolor{green!25}2.48  &\cellcolor{green!25}2.48  &4.16  & \cellcolor{red!25}5.33 &\cellcolor{orange!40} 4.66 & 2.80 & 2.92 & 2.96  &3.57  &3.36  &3.24  &3.54  &2.67  &3.43  &3.42  &3.38  &2.62  &2.59  &4.08  &4.15  \\
SOD5 &2.93  &2.92  &2.91  &3.70  &3.14  &\cellcolor{green!25}2.81  &2.86  &3.17  & \cellcolor{red!25}4.93 & 3.50 & 3.18 & 3.09 & 2.86  &\cellcolor{orange!40}4.11  &4.10  &4.00  &3.74  &3.73  &3.50  &3.45&3.41  &3.50  &3.48  &3.32  &3.44  \\
SOD6 &3.72  &3.74  &3.73  &3.96  &3.96  &3.64  &3.61  &4.04  & \cellcolor{red!25}8.14 &\cellcolor{orange!40} 4.70 & 3.85 & 3.83 & 3.96  &4.21  &4.39  &3.68  &3.73  &3.62  &3.60  &3.68  &3.65  &3.51  &\cellcolor{green!25}3.46  &3.56  &3.58  \\
SOD7 &\cellcolor{green!25}3.24  &3.25  &3.26  &3.46  &3.43  &3.29  &3.27  &3.46  & \cellcolor{red!25}5.61 & 4.11 & 3.58 & 3.62 & 3.77  &4.26  &\cellcolor{orange!40}4.34  &3.77  &3.79  &3.54  &3.46  &3.70  &3.68  &3.45  &3.35  &3.44  &3.46  \\
SOD8 & 3.74 & 3.76 & 3.75 & 3.96 & 3.98 &3.64 &3.64 & 4.15 & \cellcolor{red!25}9.07 & 4.42 & 3.99 & 4.12 & 4.27  &\cellcolor{orange!40}4.46  &4.26  &4.18  &4.05  &3.62  &3.96  &3.99  &3.90  &3.49  &\cellcolor{green!25}3.45  &3.70  &3.73  \\
SOD9 & \cellcolor{green!25}3.61 & 3.63 & 3.65 & 3.98 & 3.86 & 3.73 & 3.62 & 4.08 & \cellcolor{red!25}7.69 & 4.29 & 3.76 & 3.76 &4.09  &4.54  &4.34  &4.27  &4.47  &3.88  &\cellcolor{orange!40}4.55  &4.30  &4.21  &3.81  &3.71  &4.38  &4.38  \\
\midrule
Average   &4.54  &4.50  &4.46  &4.40  &4.36  &4.45  &4.44  &4.39  &\cellcolor{red!25}13.07  &4.73  &4.96  &4.38  &4.41  &5.49  &\cellcolor{orange!40}5.94  &4.69  &4.70  &4.58  &4.63  &4.50  &4.41  &4.33  &\cellcolor{green!25}4.23  &4.37  &4.50  \\
Rank   &13  &12  &11  &7  &3  &10  &9  &6  &\cellcolor{red!25}22  &18  &19  &5  &8 &20  &\cellcolor{orange!40}21  &16  &17  &14  &15  &12  &8  &2  &\cellcolor{green!25}1  &4  &12  \\
\bottomrule
\end{tabular}%
}
\begin{center}
\parbox{0.9\textwidth}{%
Background meaning: 
\colorbox{green!25}{\hspace{1em}} lowest error \quad
\colorbox{orange!40}{\hspace{1em}} second highest error \quad
\colorbox{red!25}{\hspace{1em}} highest error
}
\end{center}
\end{table}

None of the tested $k$NN variants clearly outperforms the remaining variants and the mean errors for methods $M_{1-13}$ are similar to those reported in~\cite{Sospedra2023}. Small discrepancies in the mean errors reported in Table~\ref{tab:3Derrors} and~\cite{Sospedra2023} could be explained by differences in hardware and software used in both studies. Except for $M_9$, all $k$NN variants yield similar average 3D positioning errors with $M_{1-8}$ and $M_{12-13}$ differing at most 18\,cm (4.1\% of the lowest average 3D positioning error) and $M_{10-11}$ yielding only 8.4\% and 13.8\% higher average errors than the best performing variant $M_5$.

For the FRNN only the variant using Cosine distance ($M_{16}$) yields competitive performance, with an average 3D positioning error 7.6\% above the average error of $M_5$. Euclidean and Cityblock distances yield 25.9\% and 36.2\% higher errors than $M_5$. At the same time, Cosine achieves the highest average coverage ratio of the three tested FRNN variants. The proposed ARNN is less sensitive to the distance metric. The average 3D positioning errors with Euclidean, Cityblock, and Cosine distances are within 12\,cm
and between 5.0\% and 7.8\% larger than for $M_5$. Thus, their performance is on a similar level than most $k$NN variants. Besides better performance, the ARNN variants also yield coverage ratios (see Table~\ref{tab:CRs}) that are 2 to 4 percentage points higher than those of the FRNN variants.

\begin{table}[t]
\caption{Coverage ratios of RNN implementations.}
\label{tab:CRs}
\centering
\scriptsize
\resizebox{\columnwidth}{!}{%
\begin{tabular}{l|*{3}{r}|*{3}{r}|*{6}{r}}
\toprule
Dataset & $M_{14}$ & $M_{15}$ & $M_{16}$ & $M_{17}$ & $M_{18}$ & $M_{19}$ & $M_{20}$ & $M_{21}$ & $M_{22}$ & $M_{23}$ & $M_{24}$ & $M_{25}$ \\
\midrule
DSI1  & 93.39 & 90.52 & 99.14 & 98.28 & 90.80 & 100.00 & 98.28 & 99.14 & 99.14 & 99.14 & 100.00 & 100.00 \\
DSI2  & 93.39 & 90.52 & 99.14 & 98.28 & 99.14 & 98.85 & 98.28 & 99.14 & 99.14 & 99.14 & 100.00 & 100.00 \\
LIB1  & 94.23 & 90.16 & 91.73 & 99.39 & 98.30 & 100.00 & 99.39 & 99.58 & 98.56 & 98.94 & 100.00 & 100.00 \\
LIB2  & 93.46 & 90.80 & 92.34 & 90.06 & 97.18 & 100.00 & 90.06 & 90.06 & 97.18 & 97.31 & 100.00 & 100.00 \\
TUT1  & 91.63 & 90.41 & 90.41 & 90.20 & 90.41 & 90.61 & 90.20 & 90.20 & 90.41 & 90.41 & 99.39 & 98.57 \\
TUT2  & 98.86 & 90.91 & 97.73 & 93.75 & 98.30 & 99.43 & 100.00 & 100.00 & 100.00 & 100.00 & 99.43 & 99.43 \\
TUT3  & 90.13 & 90.13 & 90.48 & 90.43 & 90.81 & 90.18 & 90.43 & 90.43 & 90.81 & 90.81 & 90.18 & 90.18 \\
TUT4  & 90.82 & 90.53 & 90.67 & 90.10 & 92.68 & 91.68 & 90.10 & 90.10 & 92.68 & 92.68 & 91.68 & 91.68 \\
TUT5  & 90.02 & 90.63 & 92.26 & 96.95 & 90.84 & 90.22 & 97.25 & 96.95 & 97.96 & 97.96 & 90.22 & 90.22 \\
TUT6  & 90.81 & 90.26 & 90.06 & 92.60 & 93.90 & 92.08 & 92.60 & 92.60 & 93.90 & 93.90 & 92.08 & 92.08 \\
TUT7  & 90.78 & 90.24 & 90.03 & 95.68 & 96.18 & 95.01 & 95.68 & 95.68 & 96.18 & 96.18 & 95.01 & 95.01 \\
UJI1  & 90.91 & 90.28 & 90.37 & 91.36 & 90.55 & 92.26 & 91.36 & 91.36 & 90.55 & 90.55 & 92.26 & 92.26 \\
UJI2  & 90.83 & 90.33 & 90.06 & 90.60 & 90.31 & 93.82 & 90.60 & 90.60 & 90.31 & 90.31 & 93.82 & 93.82 \\
SOD01 & 91.43 & 90.71 & 90.60 & 90.36 & 90.48 & 90.36 & 90.36 & 90.36 & 90.48 & 90.48 & 90.36 & 90.36 \\
SOD2  & 90.93 & 90.58 & 90.35 & 95.58 & 97.67 & 91.40 & 95.58 & 96.74 & 97.67 & 98.37 & 91.40 & 91.40 \\
SOD3  & 90.12 & 90.12 & 90.12 & 99.77 & 95.93 & 98.95 & 99.77 & 99.77 & 95.93 & 95.93 & 98.95 & 98.95 \\
SOD4  & 90.70 & 90.81 & 90.47 & 94.42 & 93.02 & 99.77 & 94.42 & 95.58 & 93.02 & 93.02 & 99.77 & 99.77 \\
SOD5  & 92.21 & 90.23 & 91.40 & 90.70 & 90.58 & 90.23 & 90.70 & 90.70 & 90.58 & 90.58 & 90.23 & 90.23 \\
SOD6  & 95.10 & 90.69 & 92.55 & 99.02 & 100.00 & 100.00 & 99.02 & 99.02 & 100.00 & 100.00 & 100.00 & 100.00 \\
SOD7  & 94.12 & 94.12 & 90.59 & 90.69 & 94.02 & 90.30 & 90.69 & 90.69 & 94.02 & 94.02 & 90.20 & 90.20 \\
SOD8  & 90.49 & 90.49 & 96.08 & 98.14 & 100.00 & 98.04 & 99.12 & 99.12 & 100.00 & 100.00 & 98.04 & 98.04 \\
SOD9  & 93.43 & 90.49 & 95.10 & 96.08 & 97.06 & 96.85 & 100.00 & 100.00 & 97.06 & 100.00 & 96.86 & 96.86 \\
\midrule
Average & 92.17 & 90.63 & 92.35 & 94.20 & 94.46 & 95.00 & 94.72 & 94.90 & 95.25 & 95.44 & 95.45 & 95.41 \\
\bottomrule
\end{tabular}%
}
\end{table}
WARNN variant $M_{23}$ yields the lowest average 3D positioning error of all 25 tested algorithms, with an average error 3.0\% smaller than the lowest error of any $k$NN variant. In addition, the WARNN seems to be insensitive to the choice of the distance metric and weighting function. The errors of all six tested variants were within 28\,cm, with the worst performing variant ($M_{25}$) yielding an average error 3.4\% larger than the error of $M_5$. In addition, all WARNN variants outperformed all FRNN and ARNN variants and yielded coverage ratios of approximately 95\% (all within 0.72 percentage points).

\subsection{Test with varying error threshold}

For the tests in the previous subsection an error threshold of $\tau_{\epsilon}=5$\,m was used. This value was chosen because almost all $k$NN variants yielded average 3D positioning errors below 5\,m in our tests and in~\cite{Sospedra2023}. However, modifying the error threshold $\tau_{\epsilon}$ in the training phase could theoretically affect both accuracy and coverage ratio. We, therefore, studied the impact of varying $\tau_{\epsilon}$ for the best performing RNN variant $M_{23}$.

\begin{figure}[t]
\centering
\includegraphics[width=.75\textwidth,clip=true, trim=1cm 0.25cm 0.5cm 1.0cm]{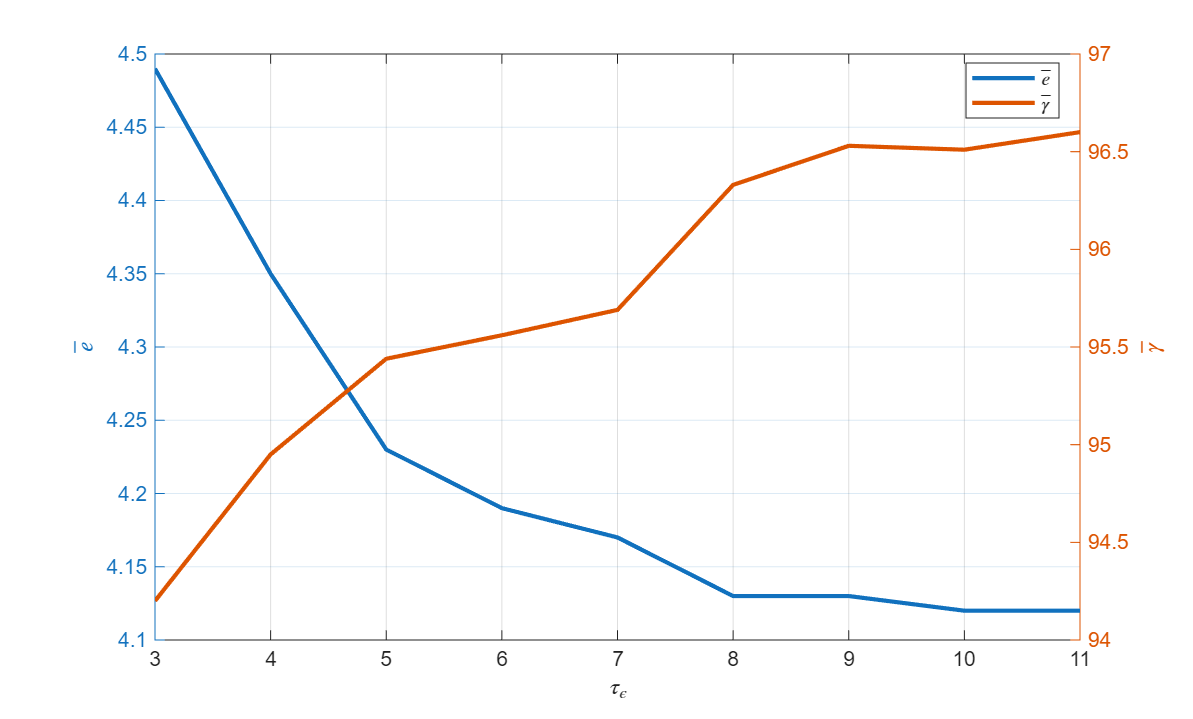}
\caption{
3D average positioning errors and corresponding coverage ratios of $M_{23}$ for varying $\tau_{\epsilon}$.
}
\label{fig:tau}
\end{figure}

Figure~\ref{fig:tau} shows the 3D average positioning errors and corresponding coverage ratios for threshold values $\tau_{\epsilon}=\{3,4,\ldots,11\}$\,m. Values below 3\,m are not reported as they did not ensure a coverage ratio of at least 90\% for some datasets. Unsurprisingly, the coverage ratio increased with larger $\tau_{\epsilon}$ because fewer training samples had $r_i=0$, resulting in more training samples being available during the test phase. More surprisingly, increasing the error threshold in the training phase decreased the 3D average positioning error of $M_{23}$ by up to 11\,cm compared to using $\tau_{\epsilon}=5$\,m, which is a reduction of 2.6\%. Once $\tau_{\epsilon} \geq 8$\,m the average error seemed to stabilize.

If individual threshold values $\tau_{\epsilon}$ yielding the mean 3D positioning errors would be chosen for all dataset, the average 3D positioning error could be further reduced to 4.05\,m with a coverage ratio of 95.81\%. This suggests that using $\tau_{\epsilon}$ as a parameter, which should be optimized for each dataset individually, could be beneficial.


\section{Conclusions}\label{sec:Disc}

This paper proposed two modifications of the Fixed Radius Near Neighbor search with adaptive radii and compared their performance with the performances of the FRNN as well as 13 $k$ Nearest Neighbors variants for indoor positioning on 22 WiFi fingerprint datasets. This extensive comparison was motivated by~\cite{Sospedra2023}, which showed that most $k$NN variants yield similarly positioning errors when evaluated over a large number of datasets, although for single datasets considerable differences in the positioning errors could be observed.

Our study confirmed these results for the 13 tested $k$NN variants and showed that, when assuming a coverage ratio of at least 90\%, the traditional FRNN performs worse than most $k$NN variants. One reason could be the use of fixed radii in the FRNN, which theoretically works well for equally spaced samples but potentially causes performance degradation for unequally distributed samples. Hence, we proposed the Adaptive Radius Near Neighbor search, which takes into account the sample distribution around each training sample when determining adaptive radii. Using adaptive radii reduced the positioning errors to a level achieved my most $k$NN variants.

One shortcoming of the ARNN is, however, that it ignores the varying radii associated with different training samples when computing position estimates for test samples from the locations of these training samples. Therefore, we proposed an extension of the ARNN that weights the contributions of all training samples to the final estimate. We propose two variants of this weighted ARNN; one with a commonly used inverse distance weighting function and one with a custom adaptive decay factor. Based on the results of our extensive evaluation the use of weighting mechanisms enables average positioning errors close or even lower than those of the best $k$NN variants. We finally showed that the accuracy of our WARNN could be further improved by optimizing the error threshold value $\tau_{\epsilon}$ used in the training phase.

It has to be noted that, unlike the traditional $k$NN and FRNN, both ARNN and WARNN incur computational cost in the training phase. Future research should, therefore, investigate potential methods for reducing this cost. Furthermore, the performance of ARNN and WARNN variants for other regression tasks as well as classification tasks should be tested and all algorithm hyperparameters should be optimized. Finally, the performance of ARNN and WARNN variants in scenarios where test samples originate from classes absent in the training set should be investigated, as under such conditions radius-based nearest neighbor search is theoretically expected to be the better choice than $k$NN-type methods due to its possibility to return no estimate when no training samples sufficiently similar to the test sample could be identified.


\section*{Data Usage and Reproducibility}

The data and code necessary to reproduce the results presented in this article will be made publicly available after acceptance by the IEEE.

\vspace{12pt}
\end{document}